# Comparing several heuristics for a packing problem

Camelia-M. Pintea[a], Cristian Pascan[b], Mara Hajdu-Macelaru[a]

[a]*Tech Univ Cluj Napoca North Univ Center 430122 Baia Mare, Romania*
[b]*SoftVision, Baia Mare, Romania*

**Abstract**

Packing problems are in general NP-hard, even for simple cases. Since now there are no highly efficient algorithms available for solving packing problems. The two-dimensional bin packing problem is about packing all given rectangular items, into a minimum size rectangular bin, without overlapping. The restriction is that the items cannot be rotated. The current paper is comparing a greedy algorithm with a hybrid genetic algorithm in order to see which technique is better for the given problem. The algorithms are tested on different sizes data.

## 1. Introduction

*Genetic Algorithms* are successfully metaheuristics used to solve in general static and dynamic $NP$-hard problems. They are adaptive heuristic search algorithms premised on the evolutionary ideas of natural selection and genetic which are, in general, applied to spaces which are too large to be exhaustively searched. Genetic Algorithms were successfully used to solve several difficult classes of problems as for example timetabling problems [2], packing problems[5, 17, 20] and vehicle routing problem [13].

The present paper will show a hybridization of *Genetic Algorithm (GA)* with a greedy technique. From a mathematical point of view, a greedy algorithm is a process that constructs a set of objects from the smallest possible constituent parts. The solution of a particular problem depends on solutions to smaller instances of the same problem. The advantage of greedy algorithm is that solutions for smaller instances are easy to understand. On the other hand, the disadvantage could be that the optimal local solutions may not lead to a good global solution.

The packing problem tested with the hybrid *GA* approach is the two dimensional Bin-Packing problem. The bin-packing problem (*2D-BPP*) is concerned with packing different sized objects into fixed sized, in general for two-dimensional bins, using as few bins as possible. The two-dimensional bin-packing problem (*2D-BPP*) and its multi-dimensional variants have many practical applications as packing objects in boxes, stock cutting, filling up containers, loading trucks with weight capacity, multiprocessor scheduling, production planning.

Some applications of *2D-BPP* are in stock cutting examples. In stock cutting, quantities of material are produced in standard sized, in general rectangular sheets. Demands for pieces of the material are for rectangles of arbitrary sizes, smaller than the sheet itself. The problem is to use the minimum number of standard sized sheets in accommodating a given list of required pieces.

Even very small, particular Bin-Packing problems are known to be NP-hard [18] as is the current problem using just one bin. In order to solve the *2D-BPP* problem one could use *Integer Linear Programming*, heuristic methods as *Local Search* and enumeration methods like *Branch-and-Bound*. These methods could prove to be useful in general for small data-sets or produce very sub-optimal packing for larger data. For complex data-sets there are used approximation algorithms which run fast and produce near-optimal packing with quality guarantees.

In the present paper one of the rules involved in the greedy shelf technique of [8] is used. In the design of these algorithms is used one of the rules: decreasing width, increasing height, etc. and then greedily pack them one by one in this order. Shelf algorithms can pack the rectangles without sorting them first. There could be used some of the following rules: *Bottom-Left, Next-Fit, First-Fit, Best-Fit*. In our implementation is involved the *Bottom-Left Fill (BLF)* technique. The implementation is not difficult, having a very fast running time and a relatively good quality guarantee.

## 2. Literature review

The current section illustrates some useful techniques for solving Bin-Packing problems. *BPP* instances are usually solved with fast heuristic algorithms. One of them is *First-Fit-Decreasing (FFD)* algorithm [8]. Here, first, the items are placed in order of non-increasing weight. Then the items are picked and placed into the first, still empty, bin, which is big enough to hold them. If there is not any

*Email addresses:* cmpintea@yahoo.com (Camelia-M. Pintea), pascan_cristian@yahoo.com (Cristian Pascan), maram@ubm.ro (Mara Hajdu-Macelaru)

bin left the item can fit in, a new bin is started. Another fast heuristic is *Best-Fit-Decreasing (BFD)* algorithm [8]. The difference from *FFD* is that the items are placed in the bin that can hold them with minimum empty space.

In [11] there is an ant colony and a hybrid ant colony approach, the hybrid one has good results comparing to Martello and Toth reduction procedure *MTP* [14] that is slower for large data-sets but has very good results. The ant-based approach is generic and is based on the idea of reinforcement of good groups through binary couplings of item sizes. In order to perform, the memetic algorithm for solving *BPP* with rotations [6] is using specific evolutionary and local search operators.

In [12], in a particular case, is developed a particle swarm algorithm for the multiobjective two-dimensional bin packing problem *MO2DBPP*. It is considered in addition to the minimization in the number of bins, the minimization of the imbalance of the bins according to a center of gravity.

## 3. The statement of the problem and the greedy technique

The section starts with the mathematical statement of the two-dimensional Bin-Packing problem based on [8], followed by greedy technique.

*2D Bin-Packing Problem.* Given a finite set of rectangular boxes $E = \{e_1, e_2, ... e_n\}$ with associated sizes $W = \{(x_1, y_1), (x_2, y2), ...(x_n, y_n)\}$ such that $0 \leq x_i, y_i \leq L^*$, where $L^*$ denotes the minimal size of a bin, place without overlapping all or some of the boxes from $E$ into the rectangular bin with sizes $X \geq L^*, Y \geq L^*$ such as the empty space is reduced to minimum.

In [15] the *2D Bin-Packing Problem (2D-BPP)* is defined as packing a finite number of two-dimensional objects, squares or rectangles, into a two-dimensional bin of a given *height* and infinite *length*, minimizing the total length required. The rotations of the rectangles are not considered. The *BPP* is extended for larger dimension of bins. For example, if $n = 3$, the *3D-BPP* studies cubes or rectangular solids and the bin has a square or rectangular base and infinite height or length.

In a *greedy technique* an optimal solution is constructed in stages. At each step it should take a decision, the best possible based on given restrictions. This decision is not changed later, so each decision should assure feasibility.

## 4. BLF-based Genetic Algorithm for 2D Bin-Packing problem

*Bottom Left Fill (BLF)* technique and the hybridization of a genetic algorithm with *BLF* are further described.

*4.1. Bottom Left Fill approach for 2D-BPP*

The *Bottom Left Fill (BLF)* technique is based on shelf algorithms [8].

The *BLF* pseudo-code is further illustrated. The greedy technique for *2D-BPP* depends on the dimensions of the rectangle. The input parameters of *BLF* function are the dimensions of the rectangle $i$, *width* and *height* and the *width* of a current bin. *BLF* function is used only for one bin and returns the coordinates $(x, y)$ for the rectangle.

The first rectangle is placed at null coordinates of the bin. It is created a list of points. In general, for a position $j$, for each rectangle is chosen the closer bottom left point $(x_j, y_j)$ where it can be placed the rectangle. This point is removed from the list and are inserted two new points: $(x_j + width_j, y_j)$ and $(x_j, y_j + height_j)$. If the point is not founded in the points list, it is necessary to add to the list a point with coordinates 0 and $max(height + y)$ based on the previous rectangle. The following conditions need to be accomplished in order to decide if it can be placed a rectangle $k$ in a given position.

- Accomplish the inequality $x + width \leq width_1$, where $width_1$ is the width of the first rectangle from the bin.

- The intersection with all previous rectangles - from 1 to $k - 1$ - need to be null.

In order to prevent some of the already mentioned conditions it is better to store the points in a sorted list. *Bottom Left Fill* technique allows the representation of the solution as a permutation.

```
Procedure BLF(width, height, maxWidth)
begin
  initialize the arrays x and y
  initialize the list and add the null point
  for all rectangles
    initialize choosePoint as impossible
    while choosePoint is impossible and j < length_of_list
      if the rectangle could be placed in a specific point
        choose the point
      endif
    endwhile
    if choosePoint is possible
      update the arrays x and y
      remove the point from the position choosePoint
      from list
      add the points (x_i+width,y_i),(x_i,y_i+height)
      to the points list
    else
      if (width > maxWidth) the problem has no solution
      else x_i = 0 and y_i = max(height_k + y_k)
      where k ∈ {1, ..., i − 1}
      endif
    endif
  endfor
  solutions: the arrays x and y with (x_i, y_i)
  the coordinates of rectangle i
end
```

*4.2. Hybrid Genetic Algorithm for 2D-BPP*

As it is mentioned in the literature review from section 2, several techniques were involved for solving *2D-BPP*.

Regarding evolutionary approaches, some variants of genetic algorithms were used to solve *BPP* [5, 17, 20].

A hybrid variant referring to a simple local search inspired by Martello and Toth [14] is described in [5]. The current paper use a hybrid genetic algorithm, *GA-BLF*, based on [5] with the *BLF* technique described in section 4.1.

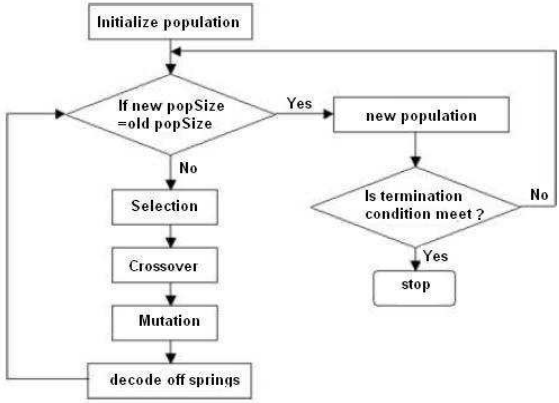

Figure 1: The illustration of successively procedures within a genetic algorithm.

Successively are described the procedures involved in the implementation of the hybrid algorithm *GA-BLF* (Figure 1). *Genetic Algorithms* are based on chromosomes representations. For the *2D-BPP* problem a chromosome is represented as a permutation of natural numbers 1 to $n$, in this case, the rectangles. $n$ is the total number of rectangles.

The hybrid algorithm *GA-BLF* starts with six initial possible solutions corresponding to the following cases. The rectangles are sorted in *ascending/descending* order of *heights*, *widths* and *surfaces areas*. For our tested data-sets, sorting the rectangles *ascending* in order of *heights* give the closest to the optimal solution. The hybrid algorithm implementation allows setting *the number of generations* and *the number of individuals per generation*. The *number of individuals* is inherited from one generation to another.

**Fitness function.** The fitness value is initialized with the height of the initial rectangle. The input parameters are chromosome widths and heights and initial rectangle width. In order to evaluate this function we need to position the rectangles with *BLF* algorithm and compute the maximal value of the height coordinate, $y$.

*Selection* function is based on elitism. At each step are chosen the best $k$ individuals to be transferred in the next generation. The *mutation* operation involves choosing a random number of pairs of genes and their interchange.

The *crossover* function return a new possible solution from two given solution. The input parameters are two chromosomes and the output is the new solution.

- Choose randomly a cutting point, $c < solution\_length$.

- It copies itself to the first $c$ positions of the second chromosome in solution.

- For each value of position $j < c$ the solution is looking for its position in the first series and swapped with the value $j$ in the same row position.

- Finally the last $n - c$ chromosomes copies itself to the solution of the first chromosome.

---
**Procedure Fitness**($widths, heights, maximum$)
begin
  order widths and heights of chromosome specifications
  calculate $x$ and $y$ arrays using **BLF** procedure
  $fit = x_1 + height_1$;
  for ($i = 2$ to $number\_of\_rectangles$)
    if ($x_i + height_i > fit$) $fit = x_i + height_i$;
    endif
  endfor
  return $fit$
end

---

## 5. Tests and Results

In order to test the hybrid genetic algorithm, *GA-BLF* quality follows several tests comparison with a greedy technique and the *BLF* technique. For this purpose are used small and large benchmarks of Hopper and Turton [22]. Data-sets dimensions are between 20 to 240 size width. The parameters used for the genetic algorithm are the number of population/generation 50, the number of good successors is 20, the surviving percentage is 30 and the maximum number of generations is considered 1000. The running time depends on the computer technical considerations: AMD 1.14 GHz obtains one second for the greedy based algorithms and for each genetic algorithm's generation.

The columns of Table 1 and Table 2 include: Hopper and Turton benchmarks [22], data-sets dimensions, the number of rectangles for each benchmark and the solutions - the empty spaces areas of the bin - for the compared algorithms: greedy, the greedy *Bottom-Left-Fill (BLF)* and the hybrid *Genetic Algorithm (GA-BLF)*. Table 1 shows the first data-set benchmarks from *20 (ht01)* to *60 (ht09)* size width. The smaller the empty space area, the better is the result illustrated in bold format.

Some visual representations for the compared algorithms are further illustrated. For the first data-set Table 1 is considered the second benchmark: *ht02* and for Table 2 is considered *c6-p1* data-set. Figure 2 shows the greedy *BLF* solution for *h02* and Figure 3 illustrates the solution for *GA-BLF*. In Figure 3 there are fewer empty spaces and the entire width (21) is smaller than for the *BLF* solution Figure 2 (24).

Table 2 shows the second set of benchmarks from *60 (c4-p1)* to *240 (c7-p3)* size width. For smaller dimensions, Table 2, shows better values, few empty spaces areas, for greedy and *GA-BLF*.

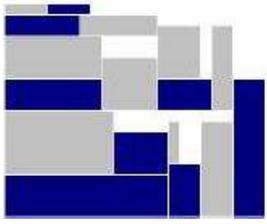

Figure 2: *BLF* solution for *ht02*.

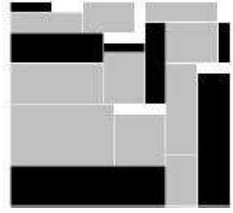

Figure 3: *GA-BLF* solution for *ht02*.

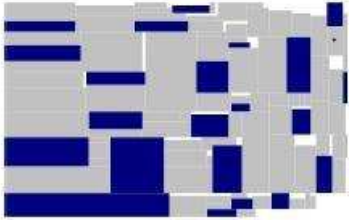

Figure 4: *BLF* solution for *c6-p1*.

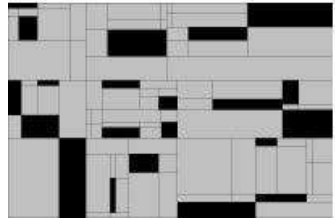

Figure 5: *GA-BLF* solution for *c6-p1*.

Table 1: Hopper and Turton benchmark data [22]

| Data-set | Dim. | No. of Rectangles | Greedy | Greedy BLF | GA BLF |
|---|---|---|---|---|---|
| ht01 | 20 | 16 | 30 | 26 | **20** |
| ht02 | 20 | 17 | 30 | 24 | **21** |
| ht03 | 20 | 16 | 29 | 23 | **20** |
| ht04 | 40 | 25 | 23 | **17** | 17 |
| ht05 | 40 | 25 | 34 | **26** | 26 |
| ht06 | 40 | 25 | 27 | **17** | 17 |
| ht07 | 60 | 28 | 46 | **32** | 32 |
| ht08 | 60 | 29 | 49 | **33** | 33 |
| ht09 | 60 | 28 | 38 | 35 | **32** |

Table 2: Hopper benchmark data [22]

| Data-set | Dim. | No. Rectangles | Greedy | BLF Greedy | GA BLF |
|---|---|---|---|---|---|
| c4-p1 | 60 | 49 | **60** | 65 | **60** |
| c4-p2 | 60 | 49 | **60** | 68 | **60** |
| c4-p3 | 60 | 49 | **60** | 65 | **60** |
| c5-p1 | 60 | 73 | **90** | 101 | **90** |
| c5-p2 | 60 | 73 | **90** | 97 | **90** |
| c5-p3 | 60 | 73 | **90** | 93 | **90** |
| c6-p1 | 80 | 97 | **120** | 127 | **120** |
| c6-p2 | 80 | 97 | **120** | 134 | **120** |
| c6-p3 | 80 | 97 | **120** | 126 | **120** |
| c7-p1 | 240 | 196 | 178 | 167 | **164** |
| c7-p2 | 240 | 197 | 190 | 166 | **163** |
| c7-p3 | 240 | 196 | 185 | **164** | 164 |

As the sizes of the data increases, for *c7-p1*, *c7-p2* and *c7-p3* the difference between the simple greedy technique and *BLF* is clearly delimited. As for the *GA-BLF*, the best results are also found for the large dimensions of Hopper benchmarks. For *c6-p1* Figure 4 illustrates the greedy *BLF* solution and Figure 5 illustrates the solution for *GA-BLF*. Figure 4 is more compact than Figure 5 and the width are 127 for *BLF* and 120 for *GA-BLF*.

Further work will focus in improving *GA-BLF* with some hybrid approach as local search, memetic techniques, involving bio-inspired methods as ant colony [4, 1, 16] and other computational intelligence tools [3, 21]. Bin-packing problems with different sizes and multi-objective variants will continue to be explored with many techniques, especially meta-heuristics, in order to solve their real life difficult applications.

## 6. Conclusion

Bin-packing problems are real life complex problems. The two dimensional approach of the problem, without considering rotations, is considered to be solve using a hybrid genetic algorithm. The genetic technique is using *Bottom-Left-Fill (BLF)*, involving iterative arrangement of rectangles from the lower left corner. Each rectangle is placed as low as possible as to not overlapping with other rectangles. The heuristic techniques are tested with good results on several real data sets.

## References

[1] Amoiralis, E.I., Georgilakis, P.S., Tsili, M.A. and Kladas, A.G. (2010) 'Ant colony optimisation solution to distribution transformer planning problem', *International Journal of Advanced Intelligence Paradigms*, Vol. 2, No. 4, pp.316–335.

[2] Burke, E.K., Elliman, D.G. and Weare, R.F. (1995) 'A hybrid genetic algorithm for highly constrained timetabling problems', *In 6th Int.Conf. on Genetic Algorithms*, pp.1258–1271.

[3] Chira, C., Dumitrescu, D. and Pintea, C.M. (2010) 'Learning sensitive stigmergic agents for solving complex problems', *Computing and Informatics*, Vol. 29, No. 3, pp.337–356.

[4] Dorigo, M., Maniezzo, V. and Colorni, A. (1996) 'The ant system: Optimization by a colony of cooperating agents', *IEEE Transactionson Systems, Man, and Cybernetics B*, Vol. 26, No. 1, pp.29–41.

[5] Falkenauer, E. (1996) 'A hybrid group in genetic algorithm for bin packing', *Journal of Heuristics*, Vol. 2, pp.5–30.

[6] Fernandez, A., Gil, C., Marquez, A.L., Banos, R., Montoya, M.G. and Alcayde, A. (2010) 'A new memetic algorithm for the two-dimensional bin-packing problem with rotations', *Distrib.Computing and Artific.Intell.*, pp.541–548.

[7] Gent, I.P. (1997) 'Heuristic solution of open bin packing problems', *Journal of Heuristics*, Vol. 3, No. 4, pp.299–304.

[8] Goodman, E.D., Tetelbaum, A.Y. and Kureichik, V.M. (1994) 'A Genetic Algorithm Approach to Compaction, Bin-Packing, and Nesting Problems', *Tech.Report* 940702, Case Center for Computer-Aided Engineering and Manufacturing, Michigan State University.

[9] Hopper, E. and Turton, B. 'A Genetic Algorithm for a 2D Industrial Packing Problem', University of Wales, Cardiff, School of Engineering, Newport Road, Cardiff, UK.

[10] Hopper, E. and Turton, B.C.H. 'An Empirical Investigation of Meta-heuristic and heuristic Algorithms for a 2D Packing Problem', University of Wales, Cardiff, School of Engineering, Newport Road, Cardiff, UK.

[11] Levine, J. and Ducatelle, F. (2004) 'Ant Colony Optimisation and Local Search for Bin Packing and Cutting Stock Problems', *Journal of the Operational Research Society, Special Issue on Local Search*, Vol. 55, No. 7, pp. 705–716.

[12] Liu, D., Tan, K., Huang, C. and Ho, W. (2008) 'On solving multiobjective bin packing problems using evolutionary particle swarm optimization', *European Journal of Operational Research*, Vol.190, No. 2, pp.357–382.

[13] Louis, S.J., Yin, X. and Yuan, Z.Y.(1999) 'Multiple vehicle routing with time windows using genetic algorithms', *In Proc.of the 1999 Congress on Evol.Comput.* pp.1804–1808.

[14] Martello, S. and Toth, P. (1990) 'Knapsack Problems, Algorithms and Computer Implementations', John Wiley and Sons Ltd., England.

[15] Pargas, R.P. and Jain, R. (1993) 'A Parallel Stochastic Optimization Algorithm for Solving 2D Bin-Packing Problems', *In Proc. 9th Conference on AI for Applications, 1993, Orlando, Florida*, pp.18–25.

[16] Pintea, C.M., Dumitrescu, D. and Pop, P.C. (2008) 'Combining Heuristics and Modifying Local Information to Guide Ant-based Search', *Carpathian J.Math.*, Vol. 24, No. 1, pp.94–103.

[17] Reeves, C. (1996) 'Hybrid genetic algorithms for bin-packing and related problems', *Annals of Operations Research*, Vol. 63, pp.371–396.

[18] Schlag, M., Liao, Y.Z. and Wong, C.K. (1983) 'An algorithm for optimal two-dimensional compaction of VLSI layouts', *Integration*, Vol. 1, No. 2,3, pp.179–209.

[19] Üçoluk, G.(2002) 'Genetic Algorithm Solution of the TSP Avoiding Special Crossover and Mutation', *Intelligent Automation and Soft Computing*, Vol. 8, pp.265.

[20] Vink, M. (1997) 'Solving combinatorial problems using evolutionary algorithms', *Technical Report Leiden University, Netherlands*.

[21] Yampolskiy, R.V., Reznik, L., Adams, M., Harlow, J. and Novikov, D. (2011) 'Resource awareness in computational intelligence', International Journal of Advanced Intelligence Paradigms, Vol. 3, pp.305–322.

[22] `http://dip.sun.ac.za/~vuuren/repositories/levelpaper/HopperAndTurtonData[1].htm`